# Multi-Label Annotation of Text Reports from Computed Tomography of the Chest, Abdomen, and Pelvis Using Deep Learning

Vincent M. D'Anniballe*,a Fakrul I. Tushar*,a,b,c Khrystyna Faryna,c,† Songyue Han,d,†, Maciej A. Mazurowskia,b, Geoffrey D. Rubina,e, Joseph Y. Loa,b

aCenter for Virtual Imaging Trials, Carl E. Ravin Advanced Imaging Laboratories, Department of Radiology, Duke University School of Medicine, 2424 Erwin Rd. Ste. 302, Durham, NC, USA, 27705

bDepartment of Electrical & Computer Engineering, Pratt School of Engineering, Duke University, Durham, NC, USA

cErasmus+ Joint Master in Medical Imaging and Applications, University of Girona, Girona, Spain

dSchool of Software Engineering, South China University of Technology, Guangzhou, Guangdong, China

eDepartment of Medical Imaging, University of Arizona College of Medicine, Tucson, Arizona, USA

*These authors contributed equally to this work

†This work was performed during an internship at Duke University.

Corresponding Author: Joseph Y. Lo (joseph.lo@duke.edu)

# Abstract

**Background:** There is progress to be made in building artificially intelligent systems to detect abnormalities that are not only accurate but can handle the true breadth of findings that radiologists encounter in body (chest, abdomen, and pelvis) Computed Tomography (CT). Currently, the major bottleneck for developing multi-disease classifiers is a lack of manually annotated data. The purpose of this work was to develop high throughput multi-label annotators for body CT reports that can be applied across a variety of abnormalities, organs, and disease states thereby mitigating the need for human annotation.

**Methods**: We used a dictionary approach to develop rule-based algorithms (RBA) for extraction of disease labels from radiology text reports. We targeted three organ systems (lungs/pleura, liver/gallbladder, kidneys/ureters) with four diseases per system based on their prevalence in our dataset. To expand the algorithms beyond pre-defined keywords, attention-guided recurrent neural networks (RNN) were trained using the RBA-extracted labels to classify reports as being positive for one or more diseases or normal for each organ system. Alternative effects on disease classification performance were evaluated using random initialization or pre-trained embedding as well as different sizes of training datasets. The RBA was tested on a subset of 2,158 manually labeled reports and performance was reported as accuracy and F-score. The RNN was tested against a test set of 48,758 reports labeled by RBA and performance was reported as area under the receiver operating characteristic curve (AUC), with 95% CIs calculated using the DeLong method.

**Results**: Manual validation of the RBA confirmed 91%–99% accuracy across the 15 different labels. Our models extracted disease labels from 261,229 radiology reports of 112,501 unique subjects. Pre-trained models outperformed random initialization across all diseases. As the training dataset size was reduced, performance was robust except for a few diseases with a relatively small number of cases. Pre-trained classification AUCs reached > 0.95 for all four disease outcomes and normality across all three organ systems.

**Conclusions**: Our label-extracting pipeline was able to encompass a variety of cases and diseases in body CT reports by generalizing beyond strict rules with exceptional accuracy. The method described can be easily adapted to enable automated labeling of hospital-scale medical data sets for training image-based disease classifiers.

# 1   Background

Machine learning algorithms have demonstrated considerable potential as disease classifiers for medical images. However, the majority of algorithms are specialized for a single organ or disease making their utility narrow in scope. This limited scope is mainly attributed to a sparsity of training data, since curating datasets for image-based classifiers has traditionally relied on radiologist annotation of the disease or its sequelae. As an alternative to image-based labeling, automated extraction of disease labels from radiology report text has the potential to address this training data scarcity and to avoid human annotation efforts (1-4).

Rule-based algorithms (RBA) are a conventional method for mining report text that utilize simple logic based on pre-defined keywords or patterns. In a landmark study, Wang et al. (5) used a RBA to extract labels of 8 thorax diseases from 108,948 chest X-ray reports to effectively train an image-based disease classifier. Using a similar method for CT, Draelos et al. (6) demonstrated the broad applicability of RBA-obtained labels by mining the more complex reports associated with over 36,000 chest CT volumes to train a classifier for 83 chest abnormalities. However, a major limitation of RBAs is that their performance and scope is reliant on the completeness of dictionaries containing pre-defined keywords. Furthermore, the radiologist's interpretation that accompanies a CT is usually composed in a free or semi-structured text form, rendering the extraction of disease labels using simple logical rules a nontrivial task (7).

To improve their utility, RBA-extracted labels can then be used to train neural networks that deviate from strict rules by learning salient semantic features, a form of natural language processing (NLP) (8, 9). For example, Steinkamp et al. (10) trained a recurrent neural network (RNN) to classify disease in pathology reports written in unseen formats, suggesting the network had learned a generalizable encoding of the semantics. Building upon this NLP approach, Yuan et al. (11) combined a pre-trained word embedding model with a deep learning-based sentence encoder to classify pulmonary nodules in a diverse set of radiology reports from different universities. While promising, it is often difficult to determine which semantic or structural features of the reports that the model perceives as most salient. To improve the interpretability of NLP-based classifiers, an attention-guided RNN (12) can be used to project the attention vector onto report text (13), allowing the user to visualize the words that a model is giving the most weight to when classifying an abnormality. For example, Banerjee et al. (14) demonstrated that an attention guided-RNN could be used to visualize synthesized information on pulmonary emboli from thoracic CT free-text radiology reports.

In this study, we propose a framework for automated, multi-disease label extraction of body (chest, abdomen, and pelvis) CT reports based on attention-guided RNNs trained on RBA extracted labels. For each organ system, a RNN was trained to classify the lungs/pleura, liver/gallbladder, kidneys/ureters as being positive for one or more of four different diseases or normal. Although there has been extensive work in radiology report labeling, to our knowledge, there are no related works that demonstrate the utility of an RBA to train deep learning-based NLP disease classifiers in such a breadth of organ systems, diseases, and body CT reports.

The main contributions of this study are threefold:
1) To develop a RBA that can meet the challenges of free-text narration in radiology CT reports.

2) To broaden the scope of our extracted labels by training attention-guided RNNs to perform multi-label disease classification of CT reports.

3) To determine alternative factors that influence disease classification performance including random vs. pre-trained embedding and different sizes of training dataset.

## 2   Materials and Methods

Institutional Review Board (IRB) approval was obtained, and informed consent was waived for this retrospective study that was compliant with the Health Insurance Portability and Accountability Act. In this section, we first describe the dataset that was used. Then, we outline the development processes of our RBAs and the subsequent addition of attention-guided RNNs to enable multi-label classification of radiology reports. Figure 1 displays the overall workflow of this paper.

This proposed work was a considerable expansion of our two previous conference proceedings manuscripts (15, 16). Our initial demonstration (15) was focused on the binary classification of organ systems as normal vs. abnormal rather than specific disease classification. While our more recent conference proceedings manuscript experimented on multi-disease annotation (16), model performance was evaluated by aggregating diseased classes into a single abnormal class. In the present study, we report disease-specific classification performance for each previous version and the final version used in this study. Compared to these previous implementations, we expanded the RBA dictionary by adding more terms, introducing wild-card entries to tackle misspellings or grammatical errors, and increased the total number of reports threefold.

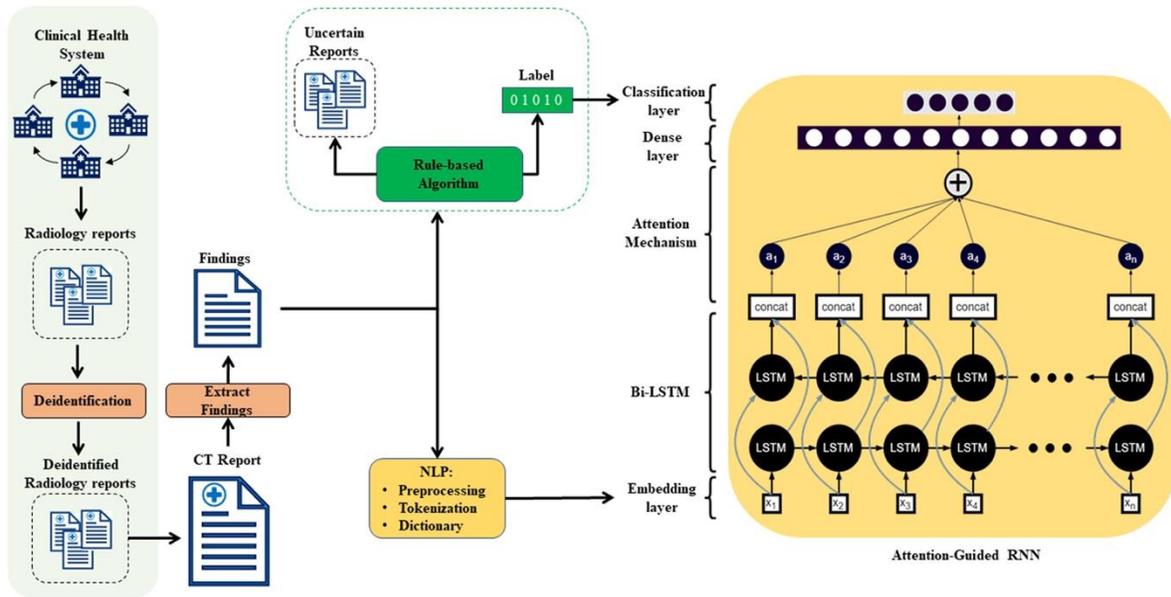

**Fig. 1 Complete workflow of this study.** Radiology reports extracted from our health system were deidentified and the findings sections were isolated. The reports were analyzed by an RBA and an attention-guided RNN to classify each report for 5 different outcomes (one or more of four disease states or normal) per organ system (lungs/pleura, liver/gallbladder, kidneys/ureters). A separate RBA and RNN was used for each organ system.

*2.1 Dataset*

A total of 261,229 chest, abdomen, pelvis structured CT reports of 112,501 unique subjects between the years 2012 to 2017 were extracted from the health system of our institution with IRB approval and deidentified. A representative example of a radiology CT report is shown in Figure 2, which contains protocol, indication, technique, findings, and impression sections. The distribution of CT protocols is shown in Figure 3.

**Protocol:** CT Chest without IV Contrast CT Abdomen and Pelvis without IV Contrast

**Indication:** C61 Malignant neoplasm of prostate (HCC), C79.51 Secondary malignant neoplasm of bone (HCC), Z00.6 Encounter for examination for normal comparison and control in clinical research program, staging

**Technique:** CT imaging of the chest, abdomen, and pelvis was performed without intravenous contrast. Coronal reformatted images were generated and reviewed. 3-D maximum intensity projection (MIP) reconstructions were performed of the chest to potentially increase study sensitivity.

**Findings:** Evaluation of the solid organs in the abdomen and pelvis is limited by the lack of IV contrast. Chest: Thyroid is unremarkable. Aortic atherosclerosis. Aorta is nonaneurysmal. Pulmonary artery is nonaneurysmal. Heart is normal in size. No pericardial effusion. Mild diffuse thickening of the thoracic esophageal wall. No axillary, mediastinal or hilar adenopathy. Coronary atherosclerosis. Small, stable nodes in the AP window. Central airways are patent. Basilar atelectasis. No focal pulmonary consolidation. No definite pulmonary nodule. Abdomen and pelvis: Liver contour is smooth. Gallbladder is unremarkable. The spleen, adrenal glands are normal. Pancreas is mildly fatty replaced. Mild bilateral perinephric stranding. Nonobstructive right renal stone in the inferior pole. No hydronephrosis. Mild stranding is seen around the distal left ureter. Aorta is nonaneurysmal. Diffuse aortoiliac atherosclerotic changes. Stomach is nondilated. Small bowel is nondilated. The appendix is normal. Bladder is unremarkable. The prostate is within normal limits of size. No free air. No free fluid. Adjacent to the distal rectum is a soft tissue nodule measuring 1.2 x 1.5 cm, unchanged from prior exam, and measures fluid density. Possibly small retroperitoneal extension of free fluid. Diffuse sclerotic osseous metastatic disease involving right femur, right and midline sacrum. Possibly L5 vertebral body. Sclerotic osseous lesions appear similar in distribution to prior exam.

**Impression:** 1. Stable osseous metastatic disease without significant interval change from September XXX, 2016. 2. No evidence of solid organ metastatic disease in the chest, abdomen or pelvis. Electronically Reviewed by: XXX, MD Electronically Reviewed on: XXX I have reviewed the images and concur with the above findings. Electronically Signed by: XXX, MD Electronically Signed on: XXX

**Fig. 2 Representative example of a body CT radiology report within our dataset.** Report consists of protocol, indication, technique, findings, and impression sections composed in a semi-structured form.

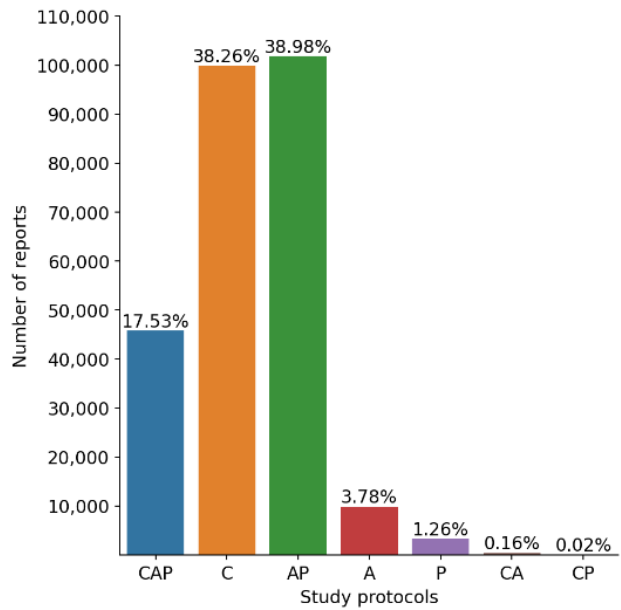

**Fig. 3 Distribution of CT protocols within our dataset.** CAP= chest, abdomen, and pelvis, C= chest, AP= abdomen-pelvis, A= abdomen, P= pelvis, CA= chest-abdomen, CP= chest-pelvis.

*2.2 Rule-Based Algorithms (RBA)*

A separate RBA was created for the lungs/pleura, liver/gallbladder, and kidneys/ureters. Each RBA was limited to the findings section of the CT reports to minimize the influence of biasing information referenced in other sections and to ensure that the automated annotation reflected image information in the current exam (e.g., indication for exam, patient history, technique factors, and comparison with priors). For example, the impression section could describe a diagnosis based on patient history that could not be made using solely image-based information. For the purpose of RNN training, reports were filtered by protocol name to ensure organ-relevant scans were used for each model. For example, only protocols that included the entire chest (CAP, C, CA, and CP) were used to train the lungs/pleura model. In order to select target disease and organ keywords for the RBA dictionary, we computed term frequency–inverse document frequency (TF-IDF) (17) on the findings sections of a random batch of 3,500 radiology reports. Informed by the prevalence of organ and disease keywords, we intentionally selected the three organ systems and four abnormal findings for each system such that the labels varied widely in location, appearance, and disease manifestations. For lungs/pleura, the four findings selected were atelectasis, nodule/mass, emphysema, and effusion. For liver/gallbladder; stone, lesion, dilation, and fatty liver. For kidneys/ureters; stone, lesion, atrophy, and cyst. A board-certified radiologist (G.D.R.) provided guidance to define the TF-IDF terms into several categories, specifically:

a) single-organ descriptors specific to each organ, e.g., pleural effusion or steatosis,
b) multi-organ descriptors applicable to numerous organs, e.g., nodule or stone,
c) negation terms indicating absence of disease, e.g., no or without,
d) qualifier terms describing confounding conditions, e.g., however, OR

e) normal terms suggesting normal anatomy in the absence of other diseases and abnormalities, e.g., unremarkable.

Appendix 1 displays the dictionary terms and their descriptor type for each organ system. Figure 4 displays an overview of the RBA's flowchart and logic. Although a separate RBA was created for each organ system, the workflow was the same. After the dictionary was refined, report text was converted to lowercase and each sentence was tokenized. In summary, the RBA was deployed on each sentence, and the number of potential diseases was counted first using the logic for the multi-organ descriptor and then the single-organ descriptor. If no potential disease labels were detected, then the normal descriptor logic was finally applied to verify normality. This process was repeated for each disease outcome allowing a report to be positive for one or more diseases or normal. Note that in this study an organ system was defined as normal not only by excluding the four diseases studied but also in the absence of dozens of abnormalities and diseases states that were not otherwise analyzed, as shown in Appendix 1. If the RBA failed to categorize the report definitively as positive for disease or normal (e.g., there was no mention of the organ system), then the report was labeled as uncertain and was not included in this study.

Upon manual review, we observed that many reports were incorrectly labeled normal due to excessively long sentences, which were either complex sentences with multiple clauses or fused together due to grammatical errors (e.g., missing periods). Such sentences were impractical to analyze with simple logic, so each report sentence was subject to a length criterion threshold for the normal outcome, another feature which made this RBA noticeably different from previous implementations.

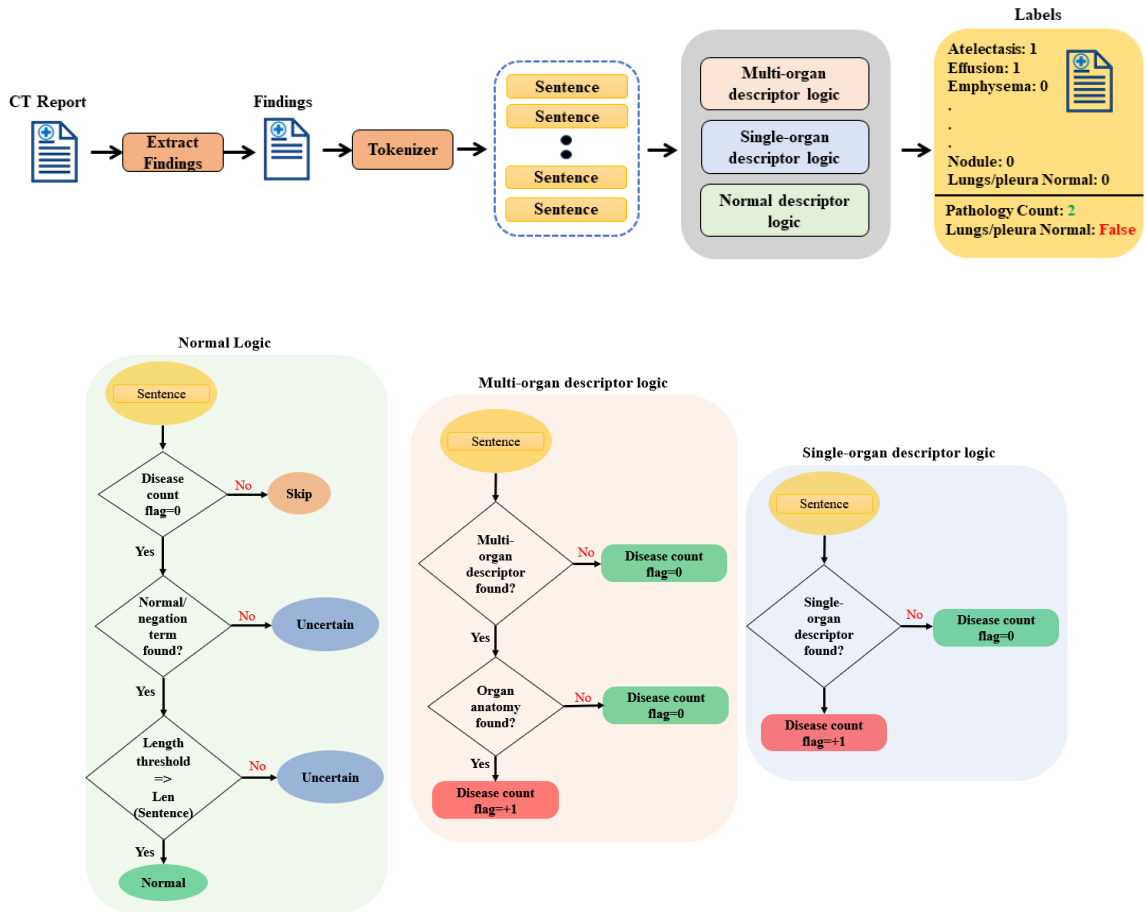

**Fig. 4 Overview of the RBAs.** (Top) The findings section of each report was extracted, then the text was converted to lowercase and each sentence was tokenized. The RBA was deployed on each sentence, and the number of diseases was counted using the multi-organ descriptor first and then the single-organ descriptor logic. If no disease labels were detected, the normal descriptor logic was applied. This process was repeated for each disease allowing a report to be positive for one or more diseases or normal for each organ system. (Bottom) The normal, multi-organ, and single organ descriptor logics.

From the full set of 261,229 reports, the lungs/pleura RBA classified a total of 165,659 reports from 74,944 subjects, the liver/gallbladder RBA classified 96,532 reports from 50,086 subjects, and the kidneys/ureters RBA classified 87,334 reports from 46,527 subjects. Note that the full set of cases does not correspond to the sum of reports for each organ system due to overlap of disease labels, where a single subject could have multiple findings across multiple organ systems. Figure 5 displays the disease distribution by organ system. Reports were randomly divided by subject into subsets for training (70%), validation (15%), and testing (15%) the RNN model.

Since the RNN depends on labels generated by the RBA, we manually validated the quality of the RBA labels. From the above test set, a test subset of 2,158 (lungs/pleura=771, liver/gallbladder=652, kidneys/ureters=749) CT reports were randomly selected, and 2,875 labels (lungs/pleura=1,154, liver/gallbladder=787, kidneys/ureters=934) were manually obtained by a Master of Biomedical Science graduate with gross anatomy training (V.M.D.) who was supervised by a board-certified radiologist (G.D.R.). This reference set was used to compare performance of the final RBA against our previous versions.

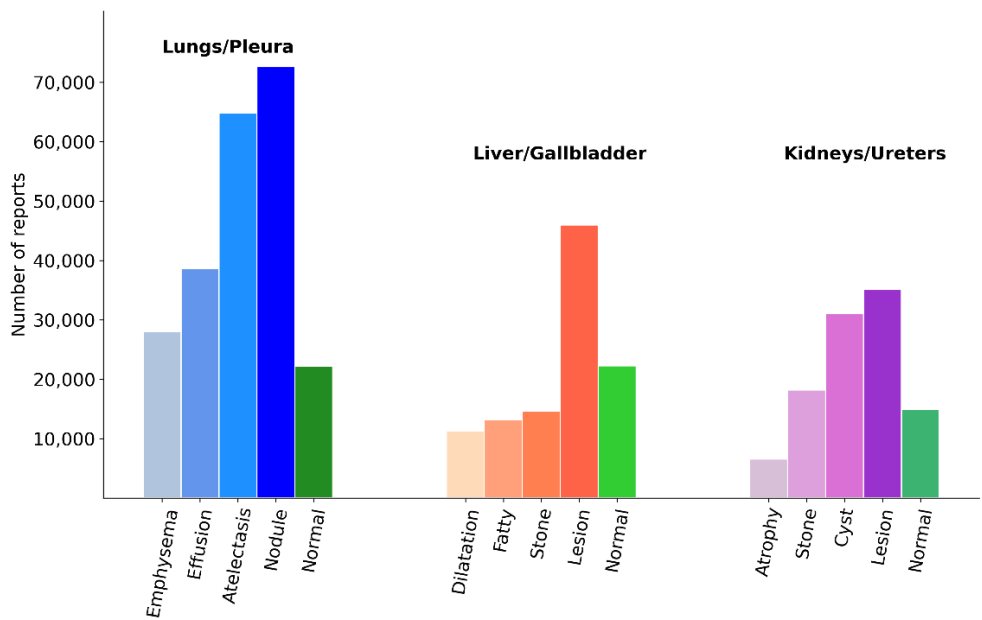

**Fig 5. Frequency of reports for each disease within our dataset.**

*2.3 Attention-Guided RNNs and Training*

A separate RNN was trained for each organ system using the corresponding RBA-annotated reports. The neural networks (Fig. 1) used in this study consisted of an embeddings layer, Bidirectional Long-Short Term Memory (BiLSTM), attention mechanism, dense layer, and final classification layer (18, 19). The BiLSTM layer was composed of 200 units and produces a sequential output. It was followed up by a 0.2 dropout layer to prevent overfitting. The attention mechanism began with a time-distributed dense layer, which received a sequential 3-dimensional input (batch size, maximum sequence length, 1), and computed the aggregation of each hidden state. Next, it was reshaped to 2-dimensional form (batch size, maximum sequence length) followed by softmax activation, which assigned weights to each hidden state to produce an attention vector. The dot product of the attention vector and sequential output of BiLSTM was the final output of the attention mechanism. It was then followed by dense and classification layers. Since outcomes for each disease were non-mutually exclusive, we used a weighted binary cross-entropy loss and modeled the outputs as independent Bernoulli distributions for each of the labels with sigmoid activation.

*2.4 Pre-Training, Datasets, Model Implementation*

In this study, we compared the multi-label classification performance of two embedding approaches: with embeddings pretrained on the PubMed+MIMIC-III (20) dataset, and without pretrained embeddings (randomly initialized embedding layer). Embeddings of 200 dimensions were used in both experiments. Afterwards, we analyzed the effect of training data size on classification performance by incrementally increasing the number of training cases from 20%, 40%, 60%, 80%, or 100% of the total dataset. To prepare the training data, a pre-processing step was applied to each report. All numbers and punctuation were removed from each "findings" section, and the text was then converted to lowercase and tokenized. The sequence of tokens was then zero padded to the length of 650 tokens per sample. The models were trained for 50 epochs using a batch size of 512. The models corresponding to the minimum of the validation loss were selected as final. In this study we used Adam optimizer and a constant learning rate of 0.0001. The models were implemented using Python TensorFlow framework. Training duration was approximately 30 minutes for each model using 2 TITAN RTX GPUs. All models' weights and code are publicly available at (https://gitlab.oit.duke.edu/railabs/LoGroup/multi-label-annotation-text-reports-body-CT ).

*2.5 Model Evaluation*

To compare our final RBA against previous versions, accuracy and F1 scores were reported using the manually obtained labels as the reference standard. Accuracy was used to assess the total correct labels for a given disease, and was calculated using true positive (TP), true negative (TN), false positive (FP), and false negative (FN) values:

$$Accuracy = \frac{TP + TN}{TP + FP + TN + FN}$$

The F1-score is defined as the harmonic mean of precision and recall (sensitivity) and was calculated as:

$$Precision = \frac{TP}{TP+FP},$$

$$Recall = \frac{TP}{TP+FN},$$

$$F_1 = 2 \cdot \frac{precision \cdot recall}{precision + recall},$$

$$False\ positive\ rate = \frac{FP}{TN+FP}$$

To compare the performance of randomly initialized versus pre-trained embeddings as well as different sizes of training data, receiver operating characteristic (ROC) area under the curve (AUC) and 95% confidence intervals (CI) were reported. The ROC curve is a plot of the false positive rate vs. the recall, and AUC is a summary metric derived from the area curve often used to report model performance. CIs were calculated using the DeLong method (21).

## 3 Results

Table 1 displays the labeling accuracy and F-score of previously reported RBAs and the final RBAs for the binary labels (present/absent) for each disease and organ system. Performance was calculated based on the manually annotated test set of 2,158 CT reports with 2,875 labels. The performance of the final RBAs were equal to or greater than both previously reported RBAs (15, 16) for all diseases, with accuracy ranging from 91% to 99% and F-score from 0.85 to 0.98.

**Table 1. Classification performance of our final RBAs compared to previously reported RBAs.**

| Organ | Label | # Pos | Han et al. (15) Acc | Han et al. (15) F-Score | Faryna et al. (16) Acc | Faryna et al. (16) F-Score | Our Final RBAs Acc | Our Final RBAs F-Score |
|---|---|---|---|---|---|---|---|---|
| Lungs/ Pleura | Atelectasis | 251 | 0.86 | 0.74 | 0.97 | 0.95 | **0.98** | **0.97** |
| | Nodule | 296 | 0.77 | 0.74 | **0.92** | **0.89** | **0.92** | **0.89** |
| | Emphysema | 193 | 0.82 | 0.45 | 0.98 | 0.96 | **0.99** | **0.98** |
| | Effusion | 205 | 0.82 | 0.53 | 0.84 | 0.58 | **0.98** | **0.97** |
| | Normal | 209 | 0.79 | 0.44 | 0.96 | 0.94 | **0.98** | **0.96** |
| Liver/ Gallbladder | Stone | 144 | 0.87 | 0.62 | 0.95 | 0.9 | **0.96** | **0.91** |
| | Lesion | 224 | 0.92 | 0.88 | 0.94 | 0.91 | **0.95** | **0.92** |
| | Dilatation | 87 | 0.86 | 0.1 | 0.9 | 0.7 | **0.98** | **0.92** |
| | Fatty | 166 | 0.97 | 0.94 | **0.98** | **0.96** | **0.98** | **0.96** |
| | Normal | 166 | 0.94 | 0.9 | 0.95 | 0.9 | **0.96** | **0.93** |
| Kidneys/ Ureters | Stone | 174 | 0.91 | 0.82 | **0.93** | **0.85** | 0.93 | 0.85 |
| | Atrophy | 94 | 0.96 | 0.85 | **0.99** | **0.97** | 0.99 | 0.97 |
| | Lesion | 238 | 0.91 | 0.87 | **0.91** | **0.86** | 0.91 | 0.86 |
| | Cyst | 234 | 0.95 | 0.92 | **0.96** | **0.94** | 0.96 | 0.94 |
| | Normal | 194 | 0.94 | 0.89 | **0.96** | **0.92** | 0.96 | 0.92 |

Comparison of classification performance between previously reported RBAs and our final RBAs using the manually annotated test set. "# Pos" is the number of positive examples for that label in the test set, Acc=Accuracy. Bolded values represent an equivalent F1-Score or increase in performance.

Table 2 displays the AUC classification performance of the attention-guided RNN with and without pre-trained embedding when applied to the test set containing 48,758 reports (23,411 reports for lungs/pleura; 13,402 reports for liver/gallbladder and 11,954 reports for kidneys/ureters). Pre-trained embedding outperformed the models trained with randomly initialized embedding for all organ systems and diseases.

**Table 2. Classification performance of randomly initialized versus pre-trained embeddings.**

| Organ | Label | # Pos | Random Initialization (AUC) | Pre-trained (AUC) |
|---|---|---|---|---|
| Lungs/ Pleura | Atelectasis | 9329 | 0.9968 (0.9961-0.9974) | **0.9973** (0.9967-0.9997) |
| | Nodule | 10183 | 0.9913 (0.9904-0.9922) | **0.9935** (0.9928-0.9943) |
| | Emphysema | 3659 | 0.9972 (0.9963-0.9982) | **0.9980** (0.9972-0.9987) |
| | Effusion | 5625 | 0.9975 (0.9970-0.9980) | **0.9984** (0.9980-0.9989) |
| | Normal | 3110 | **0.9990** (0.9985-0.9995) | **0.9990** (0.9982-0.9997) |
| Liver/ Gallbladder | Stone | 1981 | 0.7849 (0.7739-0.7059) | **0.9761** (0.9721-0.9801) |
| | Lesion | 6463 | 0.9675 (0.9646-0.9700) | **0.9946** (0.9936-0.9955) |
| | Dilatation | 1497 | 0.8120 (0.8013-0.8228) | **0.9926** (0.9906-0.9945) |
| | Fatty | 1795 | 0.9984 (0.9851-0.9917) | **0.9991** (0.9986-0.9996) |
| | Normal | 3162 | 0.9745 (0.9716-0.9773) | **0.9762** (0.9950-0.9974) |
| Kidneys/ Ureters | Stone | 2548 | 0.9562 (0.9514-0.9609) | **0.9792** (0.9764-0.9819) |
| | Atrophy | 750 | 0.9523 (0.9436-0.9611) | **0.9955** (0.9936-0.9973) |
| | Lesion | 4817 | 0.9757 (0.9731-0.9783) | **0.9900** (0.9886-0.9915) |
| | Cyst | 4164 | 0.9862 (0.9843-0.9881) | **0.9926** (0.9914-0.9939) |
| | Normal | 2048 | 0.9909 (0.9890-0.9928) | **0.9980** (0.9980-0.9992) |

Classification performance of randomly initialized versus pre-trained embeddings for each disease. "# Pos" represents the number of positives for that label. Values are reported as area under the curve (AUC) with 95% confidence interval (CI). Bolded values represent an equivalent AUC or increase in performance.

Figure 6 displays examples of the output vectors produced by the attention mechanism for each organ system. Figure 7 displays the classification performance of the attention-guided RNN with pre-trained embedding when different portions of training data were used. Figure 7(a) displays the number of reports used in the training dataset after randomly splitting in 20% increments for lungs/pleura, liver/gallbladder, kidneys/ureters. Figure 7(b) displays the classification performance after training with each increment. AUCs reached > 0.95 for all classes in each organ system when using the complete dataset in the pre-trained models. Although the performance tended to improve as more training samples were used, most labels showed a robust plateau such that performances were still within the confidence intervals for 100% of the data. The most notable drops in performance were classes with smaller sample size (e.g., stone and dilatation for liver/gallbladder and atrophy for kidneys/ureters).

### Lungs/Pleura: Nodule

Text: unremarkable thyroid no enlarged supraclavicular or axillary lymph nodes multiple enlarged paratracheal ap window and subcarinal lymph nodes are present largest of which is in the subcarinal station measuring mm short axis series image a conglomerate of left paratracheal lymph nodes measures mm short axis series image a right paratracheal lymph node measures mm short axis series image an ap window lymph node measures mm short axis series image a top normal in size right hilar lymph node measures mm short axis series image a mildly enlarged left hilar lymph node measures mm short axis heart within normal limits in size without pericardial effusion moderate coronary artery calcifications ascending aorta within normal limits in size main pulmonary artery mildly enlarged to mm series image three vessel aortic arch mild calcific atherosclerosis of the thoracic aorta trachea and proximal bronchi are clear scattered mm or less nodules in the right lung are present series image a mm nodule abuts the left hemidiaphragm in the inferior left lower lobe series image no pleural effusion or pneumothorax visualized upper abdomen demonstrates nonspecific haziness of the upper abdominal mesentery associated with multiple subcentimeter lymph nodes this is nonspecific and may be seen in setting of mesenteric panniculitis apparent tiny low density lesions in the spleen may relate to phase of contrast degenerative changes of the thoracic spine mixed indeterminate lucent and sclerotic lesions in the t and t vertebral bodies are nonspecific in appearance

### Liver/Gallbladder: Normal

Text: please see concurrent ct of the chest for findings above the diaphragm there is no evidence of focal hepatic mass or biliary obstruction the gallbladder pancreas spleen and adrenal glands are unremarkable there is no hydronephrosis multiple bilateral hypoenhancing renal lesions are too small to accurately characterize there has been significant enlargement of a periaortic lymph node measuring cm in short axis dimension previously cm series image tiny adjacent subcentimeter nodes are also increased in size there is no free fluid or free intraperitoneal gas there is no focal bowel wall thickening or evidence of intestinal obstruction scattered outside atherosclerosis in the abdominal aorta and its branches urinary bladder demonstrates no wall thickening or intraluminal filling defects there are no acute fractures or bony destructive lesions

### Kidneys/Ureters: Stone

Text: the lung bases are clear no pleural or pericardial effusion seen evaluation of the solid abdominal viscera is limited in the lack of ivcontrast the partially visualized liver gallbladder spleen pancreas and bilateral adrenal glands are normal in appearance there are bilateralnonobstructing renal calculi measuring up to mm on the left nohydronephrosis the kidneys are otherwise unremarkable in appearance there is a mm calculus at the left ureterovesicular junction series image the bowel is grossly unremarkable in appearance the urinary bladder anduterus are normal in appearance no pathologically enlarged inguinal pelvic mesenteric or retroperitoneal lymph nodes seen bone windows demonstrate no suspicious appearing osseous lesions

**Fig 6. Examples of attention vectors projected on the findings section of radiology reports.** (Top panel) a report positive for nodule in the lungs/pleura. (Middle panel) a normal report for liver/gallbladder. (Bottom panel) a report positive for stone in the kidneys/ureters. As part of standard pre-processing, all numbers and punctuation were removed and text was converted to lowercase.

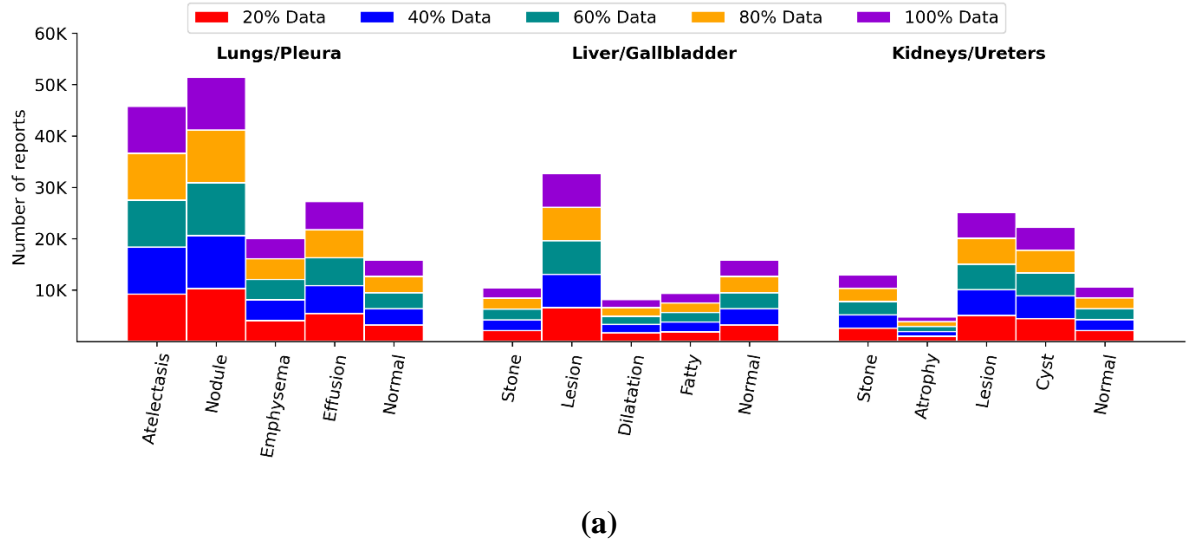

**(a)**

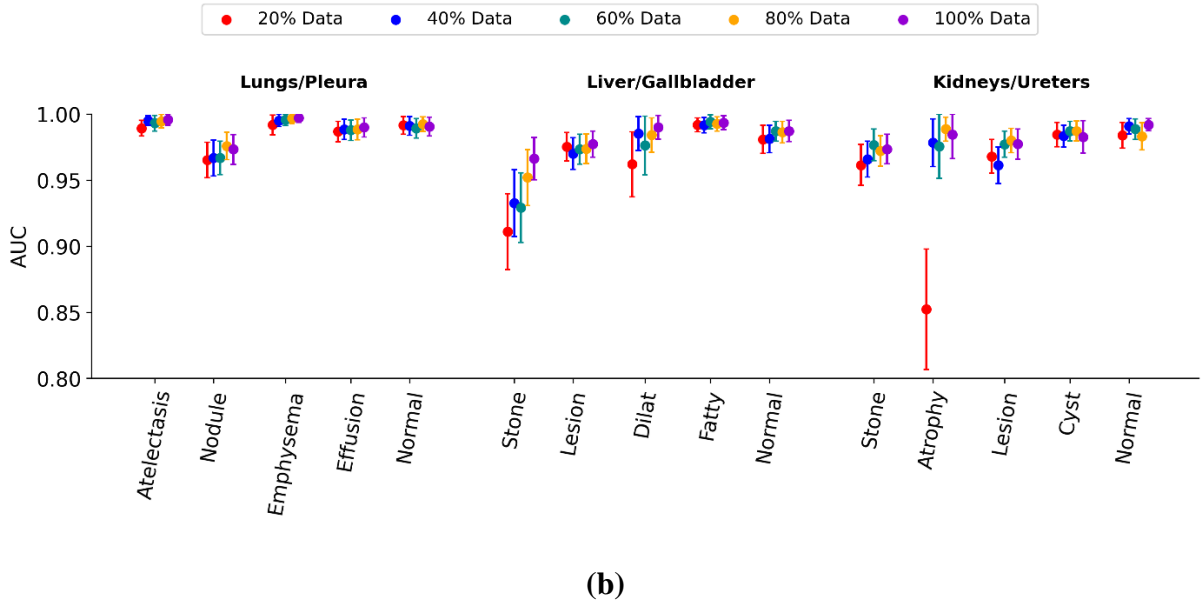

**(b)**

**Fig 7. Effect of different sizes of training data in the pretrained embedding models on classification performance.** (a) Number of reports randomly split in 20%, 40%, 60%, 80% and 100% of total training dataset for each disease by organ system. (b) Performance of models on

test-set trained with randomly split 20%, 40%, 60%, 80%, and 100% training data for each disease by organ system reported as AUC. Error bars represent 95% confidence intervals.

## 4 Discussion

Although deep learning-based disease classification algorithms have recently achieved exceptional accuracy, they often suffer from limited diversity of diseases and organ systems. This narrow scope is largely due to inadequate amounts of curated CT data where human-annotation efforts are required. As an alternative, the work described here sought to develop high-throughput, multi-disease label extractors for body CT reports that were broad in scope and could be easily adapted to new keywords and diseases. The utility of automated labeling has been demonstrated by efficiently annotating large radiology report datasets to develop image-based CT classifiers, even without specific knowledge regarding disease location (6, 22). As the foundation of our NLP algorithm, we developed RBAs that utilized simple rules to extract precise labels from radiology reports with 91-99% accuracy for all four diseases or normal across all three organ systems.

However, the RBAs alone could not provide labels for our entire dataset because radiology reports often contain variability in writing, grammar, and even variation in descriptors for the same disease between radiologists (23). To overcome this obstacle, we demonstrated that an attention-guided RNN can be trained using RBA-annotated reports to learn salient semantic features and generalize beyond simple rules or keywords to encompass more reports. Our final disease classification pipeline performed with an AUC of > 0.95 for all diseases and organ systems. Recent works investigating deep learning-based radiology report annotation have achieved similar performances, although the majority are limited to a specific disease or organ system (24). Examples include classification of pulmonary emboli in thoracic CT reports with AUC from 0.93-0.99 (14), annotation of mammography reports with a keyword accuracy of 0.96 (25), and identification of femur fractures with an F1 score of 0.97 (26).

Interpreting radiology reports often requires knowing the underlying clinical context, because language seemingly associated with disease is often used to describe normal, clinically benign findings. These common findings can account for many false positive errors when compared to manual annotations that do take into account such clinical context. Compared to previously reported RBAs (15, 16), for example, our final lungs/pleura RBA extracted labels with higher accuracy except for lung nodule. That class had many false positives for "small, calcified lung nodules," which are a common, benign finding. Similarly for liver/gallbladder, an increase in performance was seen for each class except for fatty liver disease, which had many false positives because a "small amount of fat adjacent to the falciform ligament" is also normal. Last, there was a related reason why our kidneys/ureters RBA labeling performance did not increase compared to previous work (16). Sentences describing renal diseases often contained several abnormal labels that triggered false positives for a related class but not the key finding. For example, "calcified lesion is likely a cyst" was labeled as lesion rather than cyst, and "an inferior pole left renal lesion has some calcification" was labeled as stone instead of lesion. Such difficult examples demonstrate the need for more advanced interpretation.

Further inspired by the recent wide application of deep learning-based methods in different clinical NLP tasks (10, 27-33) and effective application of word embedding (34-36), we also experimented using a multi-label disease classifier with pre-trained embedding and randomly initialized embedding layers. As expected, the attention-guided RNNs with pretrained embedding outperformed the randomly initialized models in all classes across all organ systems. Additionally, we observed that performance improved steadily with increasing number of cases. The lower frequency classes seemed to be affected greatly compared to classes having high frequency, exemplified by atrophic kidneys where performance experienced a significant drop at around 500 cases (20% of total available cases) for training.

The body CT dataset used in this study was dominated by two types of exams: chest and abdomen-pelvis CTs. In many reports, one or more of the three organ systems were out of view and not mentioned at all by the radiologist. For example, if a chest CT did not mention the kidneys, that would be labeled as uncertain by our RBA. However, in specific studies such as abdomen-pelvis CT, large organs such as the lung were often still described even if they were not completely visible e.g., "Limited view of the lung bases appear clear." This short sentence would satisfy the logic of the RBA to label the report appropriately as normal for the lungs.

There were several limitations to this study. As a general limitation of RBA techniques, it was not possible to provide disease labels for all reports within our dataset. This was often because each sentence did not satisfy the pre-defined, strict rules. To mitigate this effect, future work should expand the dictionary through discovery of new and potentially uncommon language uses. Another limitation is that, unlike when radiologists annotate images manually, the labels derived from reports tend to describe all or much of an organ system (e.g., "bibasilar atelectasis") and in some cases provide limited disease extent and location (e.g., "nodule measuring 1.8 x 2.1 cm on series 2 image 60"). Furthermore, our dataset suffered from notable class imbalance, including a low prevalence of normal cases as well as multi-fold differences between diseases, although this represented the natural prevalence within our study population. The dataset also came from a single health system, which comprises multiple hospitals but may share similarities in the reporting patterns for radiologists. Finally, this initial demonstration focused on building three separate classifiers rather than a single multi-organ model. Independent processing of diseases could have simplified the challenges imposed by multiple organ interaction, and in future work we will consider the feasibility of a single model, multi-organ approach.

## 5 Conclusions

The disease labeling pipeline described here offers numerous advantages. By using deep learning-based NLP, our algorithms were able to generalize beyond pre-defined rules and label a vast and heterogenous dataset as positive for one or more diseases or normal for three different organ systems. To the best of our knowledge, this was a first attempt in using RBA-extracted labels to train an attention-guided RNN to annotate a diverse set of diseases in a hospital-scale dataset of body CT reports. Ultimately, the work described here sought to facilitate future research in image-based disease classification algorithms by providing a general framework for labeling vast amounts of hospital-scale data in a manner that is both cost and time efficient.

## 6 Abbreviations

CT: Computed Tomography; RBA: Rule-based algorithm; RNN: Recurrent neural network; ROC: Receiver operating characteristic; AUC: Area under the curve; CI: Confidence interval; NLP: Natural language processing; IRB: Institutional review board; TF-IDF: Term frequency–inverse document frequency; CAP: Chest, abdomen, and pelvis; C: Chest; AP: Abdomen-pelvis;

A: Abdomen; P: Pelvis; CA: Chest-abdomen; CP: Chest-pelvis; BiLSTM: Bidirectional long-short term memory

# 7 Declarations

**Ethics approval and consent to participate**

This study was approved by the IRB at Duke University under protocol # Pro00082329. Informed consent was waived for this retrospective study that was compliant with the Health Insurance Portability and Accountability Act. IRB approval included permission to access the raw data. All experiments were performed in accordance with relevant guidelines and regulations.

**Consent for publication**

Not applicable.

**Availability of data and materials**

The radiology reports used in the current study cannot be shared publicly because it is impractical to ensure the removal of all protected health information from the large amount of text data. Please contact the corresponding author for data requests. The code is publicly available at https://gitlab.oit.duke.edu/railabs/LoGroup/multi-label-annotation-text-reports-body-CT.

**Competing Interests**

The authors declare that they have no competing interests.

**Funding**

The work was supported in part by seed funding from Duke Cancer Institute as part of NIH/NCI P30–CA014236, Center for Virtual Imaging Trials NIH/NIBIB P41-EB028744, MAIA Erasmus + University of Girona, and a GPU equipment grant from Nvidia Corp. These funding mechanisms were intended for general salary/equipment support and did not play any role in the study design, data collection, data analysis, data interpretation or writing of the manuscript.

**Authors' contributions**

VMD drafted the manuscript, refined the dictionary, and manually labeled the radiology reports. FIT implemented the models and performed the analyses. KF and SH helped to develop the models. MAM contributed to the development and evaluation of the study. GDR provided clinical expertise and was responsible for the overall development and editing of the manuscript. JYL contributed to the overall design, development, and writing of the manuscript. All authors have read and approved this manuscript.

**Acknowledgements**

We are grateful for helpful discussions with and data collection by Ricardo Henao PhD, Vignesh Selvakumaran MD, James Tian MD, Mark Kelly MD, Ehsan Abadi PhD, and Brian Harrawood.

**Appendix 1**. Dictionary terms used in this study.

| | Lungs/Pleura | Liver/Gallbladder | Kidneys/Ureters |
|---|---|---|---|
| **Organ Anatomy** | lung, pulmonary, lower\|upper\|middle lobe, centrilobular, perifissural, left\|right base, bases, basilar, bronch, trachea, airspace, airway | liver, hepatic, hepato, gallbladder, thegallbladder, gall bladder, biliary, bile, left\|right\|caudate\|quadrate lobe | kidney, renal, nephr, ureter, cort, medul, caliectasis, UVJ |
| **Single-organ Disease descriptors** | pneumothorax, emphysema, pneumoni, ground glass, aspiration, bronchiectasis, atelecta, embol, air trapping, pleural effusion, pneumonectomy | steatosis, cirrho, cholecystectomy, gallstone, cholelithiasis | hydronephrosis, hydroureter, nephrectomy, pelvicaliectasis, uropathy, ureterectasis, nephrolithiasis |
| **Multi-organ Disease Descriptors** | mass, opaci, calcul, stone, scar, metas, malignan, cancer, tumor, neoplasm, lithiasis, atroph, recurren, hyperenhanc, hypoenhanc, aneurysm, lesion, nodule, nodular, calcifi, opacit, effusion, resect, thromb, infect, infarct, inflam, fluid, consolidate, degenerative, dissect, collaps, fissure, edema, cyst, focus, angioma, spiculated, architectural distortion, lytic, pathologic, defect, hernia, biops, encasement, fibroid, hemorrhage, multilocul, distension, stricture, obstructi, hypodens, hyperdens, hypoattenuat, hyperattenuat, necrosis, irregular, ectasia, destructi, dilat, granuloma, enlarged, abscess, stent, fatty infiltr, stenosis, delay, carcinoma, adenoma, atrophy, hemangioma, density, surgically absent | | |
| **Negation** | no, non, other, not, none, without, rather, negative, with regards to, however is no, are no, no evidence, noevidence, limited exam for the evaluation | | |
| **Qualifiers** | acute, new, size, contour, attenuation, caliber, however, morphological | | |
| **Normal** | Normal, unremarkable, negative exam, patent, clear, no abnormalit, without abnormalit | | |